\documentclass[final]{cvpr_arxiv}

\usepackage{times}
\usepackage{epsfig}
\usepackage{graphicx}
\usepackage{amsmath}
\usepackage{amssymb}
\usepackage{ulem}
\usepackage{booktabs}
\usepackage{subcaption}
\usepackage{lipsum}
\usepackage{stfloats}
\usepackage[hang,flushmargin]{footmisc}

\usepackage[pagebackref=true,breaklinks=true,colorlinks,bookmarks=false]{hyperref}

\makeatletter
\def\thickhline{%
\noalign{\ifnum0=`}\fi\hrule \@height \thickarrayrulewidth \futurelet
\reserved@a\@xthickhline}
\def\@xthickhline{\ifx\reserved@a\thickhline
            \vskip\doublerulesep
            \vskip-\thickarrayrulewidth
            \fi
    \ifnum0=`{\fi}}
\makeatother

\newlength{\thickarrayrulewidth}
\newcommand\blfootnote[1]{%
    \begingroup
    \renewcommand\thefootnote{}\footnote{#1}%
    \addtocounter{footnote}{-1}%
    \endgroup
}

\def\cvprPaperID{2649} 
\def\confYear{CVPR 2021}
\begin{document}


\title{RefineMask: Towards High-Quality Instance Segmentation \\ with Fine-Grained Features}

\author{Gang Zhang$^{1}$ \quad Xin Lu$^{2}$ \quad Jingru Tan$^{3}$  \\ Jianmin Li$^{1}$ \quad Zhaoxiang Zhang$^{4}$ \quad Quanquan Li$^{2}$ \quad Xiaolin Hu$^{1}\footnotemark[1]$\\
$^{1}$State Key Laboratory of Intelligent Technology and Systems, Institute for AI, \\ BNRist, Department of Computer Science and Technology, Tsinghua University, \\$^{2}$SenseTime Research, $^{3}$Tongji University, $^{4}$Institute of Automation, CAS \& UCAS  \\
{\tt\small zhang-g19@mails.tsinghua.edu.cn, tanjingru120@gmail.com, zhaoxiang.zhang@ia.ac.cn}, \\
{\tt\small \{luxin,liquanquan\}@sensetime.com, \{lijianmin,xlhu\}@mail.tsinghua.edu.cn}
}

\maketitle

\blfootnote{*Corresponding Author}

\begin{abstract}
The two-stage methods for instance segmentation, \eg Mask R-CNN, have achieved excellent performance recently. However, the segmented masks are still very coarse due to the downsampling operations in both the feature pyramid and the instance-wise pooling process, especially for large objects. In this work, we propose a new method called RefineMask for high-quality instance segmentation of objects and scenes, which incorporates fine-grained features during the instance-wise segmenting process in a multi-stage manner. Through fusing more detailed information stage by stage, RefineMask is able to refine high-quality masks consistently. RefineMask succeeds in segmenting hard cases such as bent parts of objects that are over-smoothed by most previous methods and outputs accurate boundaries. Without bells and whistles, RefineMask yields significant gains of 2.6, 3.4, 3.8 AP over Mask R-CNN on COCO, LVIS, and Cityscapes benchmarks respectively at a small amount of additional computational cost. Furthermore, our single-model result outperforms the winner of the LVIS Challenge 2020 by 1.3 points on the LVIS test-dev set and establishes a new state-of-the-art. Code will be available at \href{https://github.com/zhanggang001/RefineMask/}{https://github.com/zhanggang001/RefineMask}.
\end{abstract}

\section{Introduction}

\begin{figure}[t]
    \begin{center}
    \includegraphics[width=1.0\linewidth]{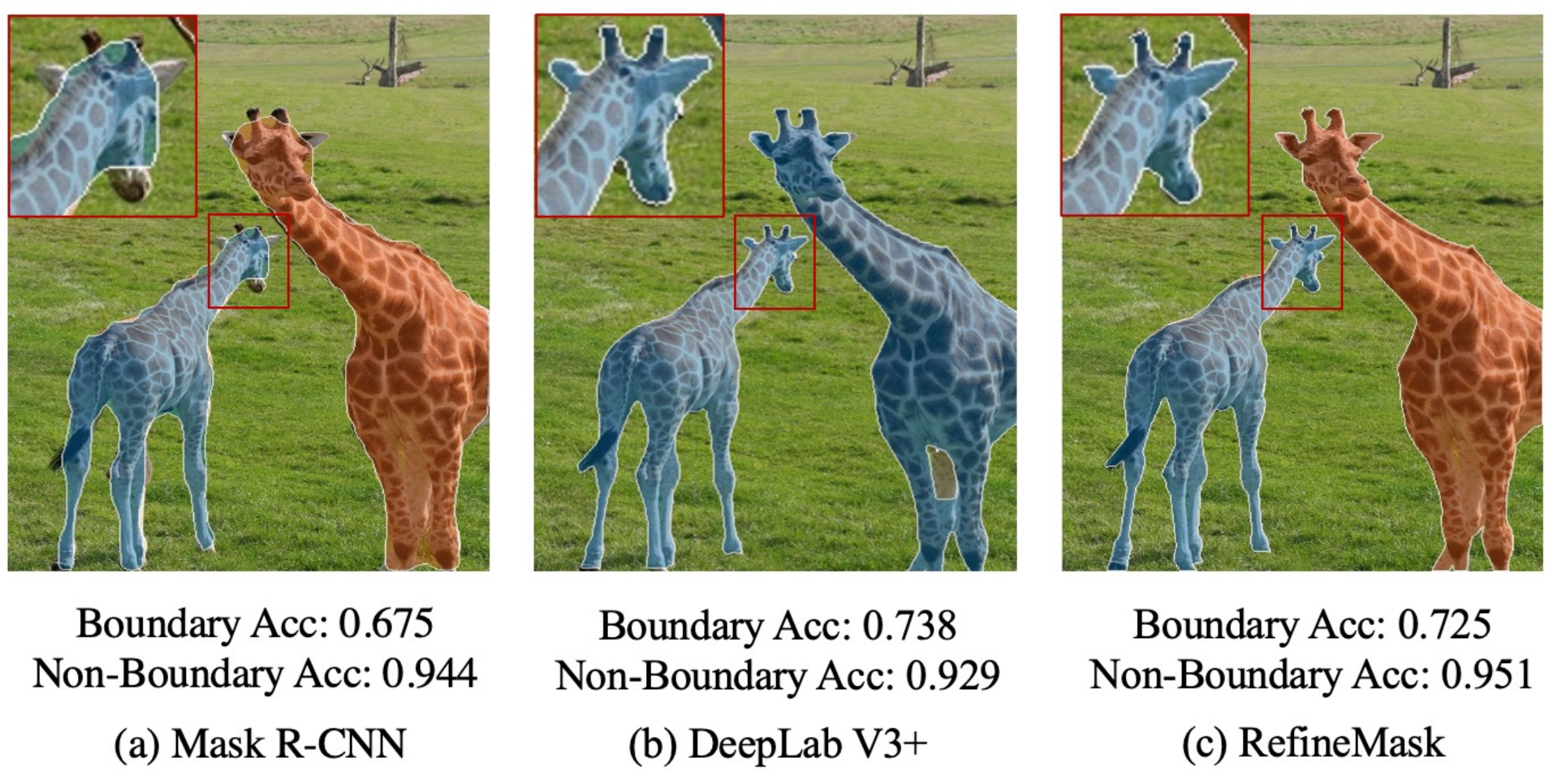}
    \end{center}
    \caption{Comparison among three methods for image segmentation: (a) Mask R-CNN, (b) DeepLabV3+, and (c) our proposed \textbf{RefineMask}. The segmentation results are indicated by the blue or orange shadows. The boundary accuracy denotes the average accuracy of the predicted masks in boundary regions, and the non-boundary accuracy is defined similarly (the definition of boundary region can be found in Section~\ref{inference_of_boundary_aware_refinement}). Both accuracies of the three methods are calculated over the entire COCO dataset~\cite{COCO}.}
    \label{fig:motivation}\vspace{-3mm}
\end{figure}

Instance segmentation~\cite{HTC,BMaskRCNN,MaskRCNN,PointRend,CondInst,SOLO} is a fundamental but challenging task in computer vision, which aims to assign each pixel into a specific semantic category and differentiate instances in the same category. The recent top-performing methods~\cite{HTC,BMaskRCNN,MaskRCNN,PointRend} for instance segmentation generally follow a two-stage paradigm. Taking Mask R-CNN~\cite{MaskRCNN} as an example, a powerful detector is first employed to generate high-quality bounding boxes and then a parallel segmentation branch is introduced to predict binary mask for each instance inside the bounding box. Specifically, in the latter step, a pooling operation, \eg, RoIAlign~\cite{MaskRCNN}, is used to extract instance features from the feature pyramid~\cite{FPN}, then pixel-wise classification is performed based on the output features of the mask head.

Despite the strong abilities provided by the powerful object detector~\cite{FasterRCNN} to locate and distinguish instances, Mask R-CNN loses the image details which are indispensable for high-quality instance segmentation task, see Figure~\ref{fig:motivation} (a). The loss of details is mainly due to two factors. Firstly, the features fed into the pooling operation are from multiple levels of the feature pyramid, and the higher-level features usually result in coarser spatial resolution. For these high-level features, it is hard to preserve details when mapping the mask prediction back to input space. Secondly, the pooling operation itself further reduces the spatial size of the features into a smaller size, \eg 7$\times$7 or 14$\times$14, which also causes information loss.

In contrast to instance segmentation, the goal of semantic segmentation is to classify each pixel into a fixed set of categories without differentiating object instances. Since semantic segmentation does not need the extreme high-level features to distinguish large instances, it can make full use of the high-resolution features, \eg P2 in the feature pyramid. Many recently proposed semantic segmentation methods~\cite{DeepLab,DeepLabV3plus,PanopticFPN,HRNet} take advantages of high-resolution features to generate high-quality semantic representation and successfully segment sharp object boundaries. These methods have higher prediction accuracy on boundary regions of objects than the two-stage instance segmentation methods, as shown in Figure~\ref{fig:motivation} (b). Moreover, it is obvious that there is no necessity to utilize any instance-wise pooling operation, \eg RoIAlign, to extract instance features in semantic segmentation, further alleviating loss of details.

Our main idea in this work is to perform instance segmentation by keeping the strong ability of current two-stage methods for distinguishing instances and supplementing the lost details with fine-grained features during the instance-wise segmenting process. To achieve this goal, we propose a new framework named RefineMask. RefineMask builds a new semantic head on the highest resolution feature map in the feature pyramid to generate fine-grained semantic features. These fine-grained features are used to supplement the lost details in the instance-wise segmenting process. Different from  Mask R-CNN, RefineMask uses a multi-stage refinement strategy in the mask head. Specifically, after the RoI-Align operation, it gradually up-scales the prediction size and incorporates the fine-grained features to alleviate the loss of details for high-quality instance mask prediction. Moreover, RefineMask uses a boundary-aware refinement strategy to focus on the boundary regions for predicting more accurate boundaries. Through fusing more fine-grained features iteratively and focusing on the boundary regions explicitly, RefineMask is able to consistently refine higher quality masks. As shown in Figure~\ref{fig:motivation} (c), RefineMask outputs much higher quality segmentation results than Mask R-CNN and obtains comparable details as state-of-the-art semantic segmentation methods, especially in hard regions such as object boundaries.

We evaluated our method on different datasets for instance segmentation and achieved significant improvements consistently. Without bells and whistles, RefineMask outperformed Mask R-CNN by 2.6, 3.4 and 3.8 points on COCO, LVIS and Cityscapes validation sets respectively.
Evaluated under the finer LVIS annotations, RefineMask trained on COCO achieved 4.1 points AP improvements over Mask R-CNN, and the gain reached 6.0 points for large objects. Furthermore, our single-model result surpassed the winner of the LVIS Challenge 2020~\cite{LVIS2020Winner} by 1.3 points on
the LVIS test-dev set and established a new state-of-the-art.

\begin{figure*}[ht]
    \begin{center}
    \includegraphics[scale=0.5]{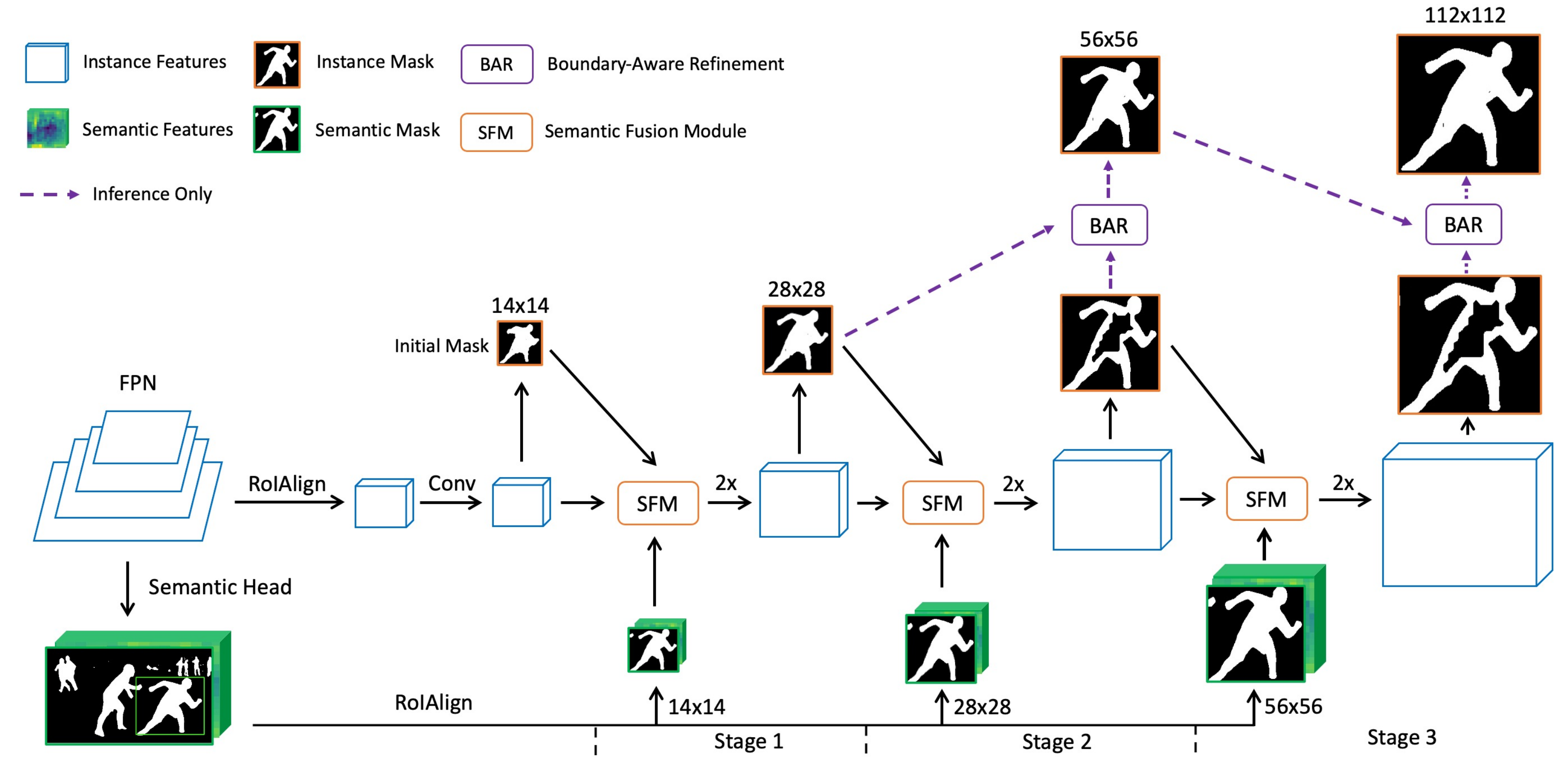}
    \end{center}
    \caption{{\bf{Framework of RefineMask.}} Based on FPN~\cite{FPN}, a mask head (the middle row) parallel to the detection head (omitted for clarity) is introduced to perform instance segmentation in a multi-stage manner, and a semantic head is attached to P2 to generate fine-grained features. Each stage has a Semantic Fusion Module (SFM) to fuse the instance features obtained from its preceding stage and the semantic features pooled from the output of the semantic head, also receiving the instance mask and the semantic mask as a guide. Moreover, a boundary-aware refinement (BAR) strategy is proposed to focus on the boundary regions for predicting more accurate boundaries in later stages.}
    \label{fig:framework}\vspace{-1mm}
\end{figure*}

\section{Related Work}
\noindent\textbf{Instance segmentation.}
The dominant methods for instance segmentation~\cite{HTC,BMaskRCNN,MaskRCNN,PointRend} often utilize a powerful detector to generate bounding boxes and then categorize each pixel inside the bounding box as a foreground or background pixel. However, as these methods rely on a pooling operation, \eg RoIAlign~\cite{MaskRCNN}, to extract canonical-size features from the feature pyramid~\cite{FPN} for each instance, which loses many details, it's hard for the segmenters to predict high quality instance masks, especially for large objects. In this work, we perform instance segmentation by incorporating information extracted from the fine-grained features stage by stage to supplement the lost details.\vspace{1mm}

\noindent\textbf{Multi-stage refinement.}
Multi-stage refinement is widely used to improve performance in object detection ~\cite{CascadeRPN,CascadeRCNN} and image segmentation~\cite{HTC,PointRend,Not_All_Pixels_Are_Equal,SharpMask}. Cascade R-CNN~\cite{CascadeRCNN} uses a sequence of detectors to regress precise bounding boxes. Deep Layer Cascade~\cite{Not_All_Pixels_Are_Equal} treats a single deep model as a cascade of several sub-models to promote semantic segmentation. HTC~\cite{HTC} designs an intertwined cascade mask head to boost performance for object detection and instance segmentation, but with too much computational cost. PointRend~\cite{PointRend} performs point-based predictions at the blurred areas iteratively for high-quality image segmentation. Different from PointRend, we refine entire objects together. SharpMask~\cite{SharpMask} shares similar motivation with us, but it focuses on the object proposal generation.\vspace{1mm}

\noindent\textbf{Methods using semantic segmentation.}
Semantic segmentation is also utilized as a supplement to instance segmentation in some recent methods~\cite{YOLACT,BlendMask,HTC,MaskLab}. HTC~\cite{HTC} introduces a semantic branch to provide spatial contexts for both object detection and instance segmentation. Mask Lab~\cite{MaskLab} combines the semantic branch with a direction branch to distinguish instances inside the bounding box. HRNet~\cite{HRNet} generates high-resolution representation via a well-designed structure, which can also benefit instance segmentation. However, all these methods still suffer from the loss of details, as they incorporate the semantic features before the instance-wise pooling operations. In this work, we supplement the lost details iteratively with the fine-grained semantic features in the refinement process.\vspace{1mm}

\noindent\textbf{Boundary-aware segmentation.}
Obtaining sharp boundaries is essential for high-quality image segmentation, and lots of work~\cite{BMaskRCNN,BAIS,PolyTransform,SegFix} make attempt on this. PolyTransform~\cite{PolyTransform} utilizes a deforming network to transform the polygons generated by the results of existing segmentation methods to better fit the object boundaries. SegFix~\cite{SegFix} presents a post-processing scheme to improve the boundary quality by replacing the originally unreliable predictions of boundary pixels with the predictions of interior pixels. BMask R-CNN~\cite{BMaskRCNN} enhances the mask features by introducing an additional contour supervision to better align with object boundaries. Unlike existing methods, we facilitate the instance boundaries through focusing on the boundary regions explicitly in later stages of the refinement process.

\section{RefineMask}
An overview of RefineMask is shown in Figure~\ref{fig:framework}. Based on the powerful detector FPN~\cite{FPN}, RefineMask relies on two small network modules, i.e. the semantic head and the mask head (the middle row, right of the FPN), to perform high-quality instance segmentation.

The semantic head takes the highest resolution feature map from the feature pyramid as input and performs semantic segmentation. The output of the semantic head keeps the same resolution as the input without using spatial compression operations such as downsampling. The fine-grained features (see the definition in Section~\ref{semantic_head}) generated by the semantic head are utilized to facilitate instance segmentation in the mask head.

The mask head performs instance segmentation in a multi-stage fashion. At each stage, the mask head incorporates the semantic features and the semantic mask extracted from the fine-grained features and increases the spatial size of the features for finer instance mask prediction. In addition to that, a boundary-aware refinement strategy is proposed in the mask head to explicitly focus on the boundary regions for predicting more crisp boundaries. We delve into details of each component in the following subsections.

\subsection{Semantic Head}\label{semantic_head}
The semantic head is a fully convolutional neural network attached to \textit{P2} (the highest resolution feature map of FPN). It consists of four convolutional layers to extract the \textbf{\textit{semantic features}} of the entire image, and a binary classifier to predict the probability of each pixel belonging to the foreground. Under the supervision of binary cross-entropy loss, it predicts a high-resolution \textbf{\textit{semantic mask}} for the entire image. We define \textbf{\textit{fine-grained features}} as the union of the semantic features and the semantic mask. These fine-grained features are further used to supplement the lost details in the mask head for high-quality mask prediction.

\subsection{Mask Head}
The mask head is a fully convolutional instance segmentation branch. In the mask head, features extracted by the 14$\times$14 RoIAlign operation are first fed into two 3$\times$3 convolutional layers to generate \textbf{\textit{instance features}}. After that, a 1$\times$1 convolutional layer is adopted to predict the \textbf{\textit{instance mask}} like Mask R-CNN but the spatial size of the mask is only 14$\times$14. This coarse mask is served as an initial mask for later refinement stages. \vspace{1mm}

\begin{figure}[t]
    \begin{center}
    \includegraphics[width=1.0\linewidth]{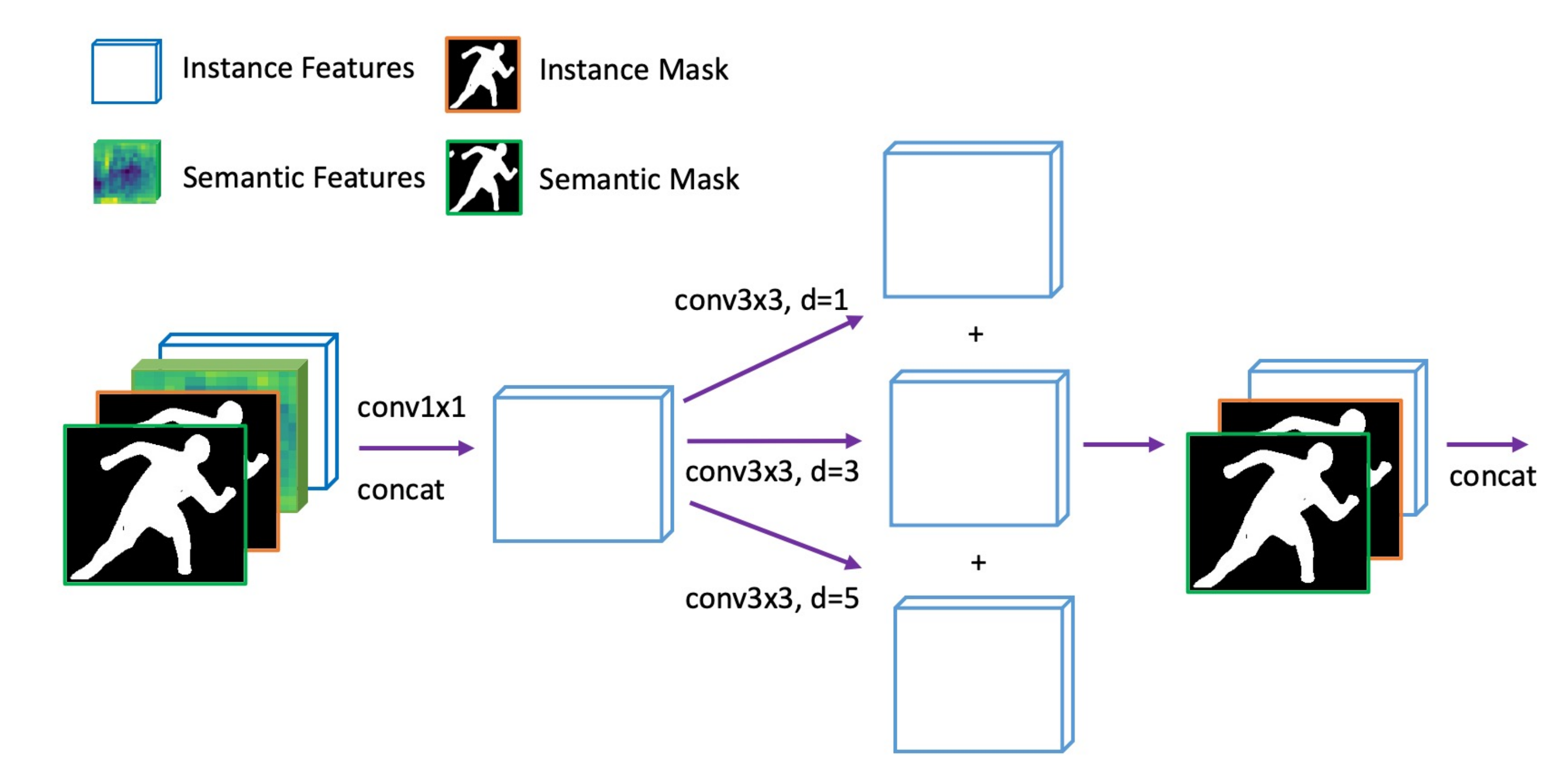}
    \end{center}
    \caption{\textbf{Semantic Fusion Module}. Four input parts are first concatenated and compressed by a 1$\times$1 convolutional layer, then three parallel convolutional layers with different dilations are used to fuse the features and masks.}
    \label{fig:fusion_module}
\end{figure}

\begin{figure}[t]
    \begin{center}
        \scalebox{0.8}{
        \includegraphics[width=1.0\linewidth]{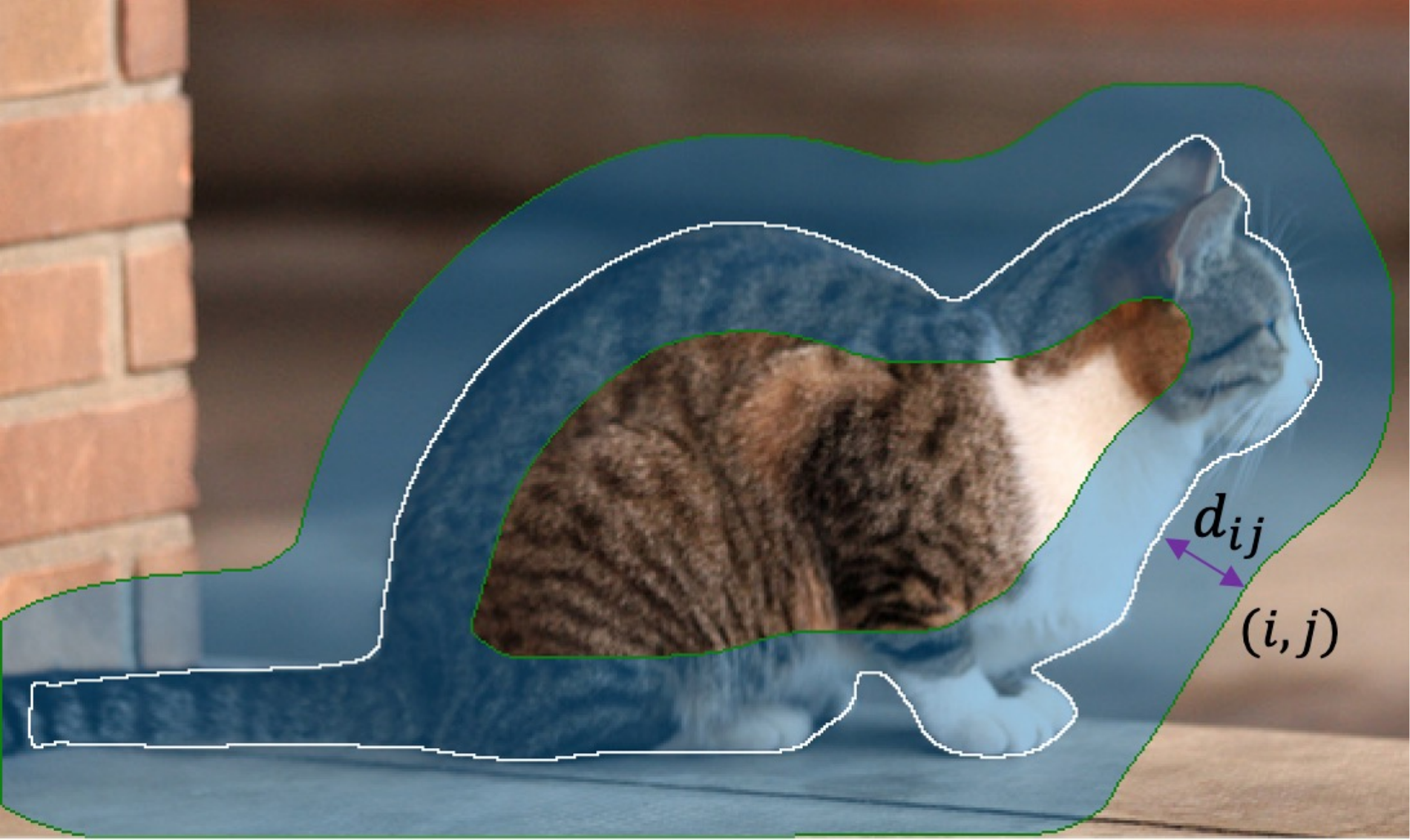}
        }
    \end{center}
    \caption{\textbf{An illustration of object boundary region}. $d_{ij}$ is defined as the Euclidean distance from pixel $p_{ij}$ to its nearest pixel on the mask contour (the white line).}
    \label{fig:boundary_example}\vspace{-2mm}
\end{figure}

\noindent\textbf{Multi-stage refinement.}\label{multi_stage_refinement} After the above process, we obtain a coarse instance mask. Next, a multi-stage refinement procedure is proposed to refine the mask in an iterative manner. Inputs of each stage are consists of four parts, \ie the instance features and the instance mask obtained from its preceding stage, the semantic features and the semantic mask pooled from the output of the semantic head. A Semantic Fusion Module (SFM) is proposed to integrate these inputs and the fused features are then up-scaled to higher spatial size (see details later). The mask head runs this refinement procedure iteratively and outputs a high-quality instance mask of resolution up to 112$\times$112.

Note that before being up-scaled to higher spatial size, the fused features in SFM are compressed with a 1$\times$1 convolutional layer to halve its channels. Therefore, although the spatial size of features grows larger and larger, the additional computational cost introduced is quite low.\vspace{1mm}

\noindent\textbf{Semantic Fusion Module.}
In order to better integrate the fine-grained features, we design a simple fusion module called \textit{Semantic Fusion Module} (\textit{SFM}) to make sure that each neuron in the mask head perceive its surrounding context, as shown in Figure~\ref{fig:fusion_module}. It concatenates four input parts of each stage mentioned above, following a 1$\times$1 convolutional layer to fuse these features and reduce the channel dimension. After that, three parallel 3$\times$3 convolutional layers with different dilations are used to fuse information around a single neuron while keeping the local details. Finally, the instance mask and the semantic mask are concatenated with the fused features again, as a guide for later prediction.

\subsection{Boundary-Aware Refinement} \label{boudare_aware_refinement}
For the purpose of predicting accurate boundaries, we propose a boundary-aware refinement strategy to focus on the boundary regions.\vspace{1mm}

\noindent\textbf{Definition of boundary region.}
Let $M^k$ denotes the binary instance mask of stage $k$, and the spatial size of the mask can be formulated as $14\cdot2^k \times 14\cdot2^k$, where $k$=1, 2, 3 (Figure~\ref{fig:framework}). The boundary region of $M^k$ is defined as the region consisting of pixels whose distance to the mask contour is less than $\hat{d}$ pixels. We introduce a binary mask $B^k$ to represent the boundary region of $M^k$, and $B^k$ can be formulated
as follows:
\begin{equation}\label{boundary_definition}
    B^k(i,j) = \left\{
            \begin{array}{lr}
            1, & \text{if } d_{ij} \le \hat{d} \\
            0, & \text{otherwise.}\\
            \end{array}
        \right.
\end{equation}
where $(i, j)$ denotes position of the pixel $p_{ij}$ in $M^k$, and $d_{ij}$ is the Euclidean distance from pixel $p_{ij}$ to its nearest pixel on the mask contour. An illustration is shown in Figure~\ref{fig:boundary_example}. For efficient implementation, we design a convolutional operator to approximate the calculation of boundary regions (details can be found in Appendix). As objects have different scales, we first resize the instance mask into a fixed size, \eg 28$\times$28 in stage 1 and 56$\times$56 in stage 2, and then calculate the boundary mask. \vspace{1mm}

\noindent\textbf{Training.} The first stage predicts a complete instance mask with a size of 28$\times$28. In the two subsequent stages whose output sizes are 56$\times$56 and 112$\times$112, only certain boundary regions are trained with supervised signals. These regions are determined by both the ground-truth mask and the predicted mask of its \textit{preceding stage}:
\begin{equation}
    R^{k} = f_{\text{up}}(B_G^{k-1} \lor B_P^{k-1})
\end{equation}
where $f_{\text{up}}$ denotes the bilinear upsampling operation with scale factor of 2, $B_G^{k-1}$ denotes the boundary region of the ground-truth mask in stage $k-1$,  $B_P^{k-1}$ denotes the boundary region of the predicted mask in stage $k-1$, and $\lor$ denotes the union of above two boundary regions. The training loss $\mathcal{L}^k$ for the $k$-th stage ($k=2, 3$) with an output size of $S_k\times S_k$ can be defined as follows:


\begin{equation}\label{boundary_loss}
    \mathcal{L}^k = \frac{1}{\delta_n}\sum_{n=0}^{N-1}\sum_{i=0}^{S_k-1}\sum_{j=0}^{S_k-1} R_{nij}^k\cdot l_{nij}
\end{equation}

\begin{equation}
    \delta_n = \sum_{n=0}^{N-1}\sum_{i=0}^{S_k-1}\sum_{j=0}^{S_k-1} R_{nij}
\end{equation}

\noindent where $N$ is the number of instances, $l_{nij}$ is a binary cross-entropy loss at position $(i, j)$ for instance $n$.\vspace{1mm}

\begin{figure}[t]
    \begin{center}
        \scalebox{0.97}{
            \includegraphics[width=1.0\linewidth]{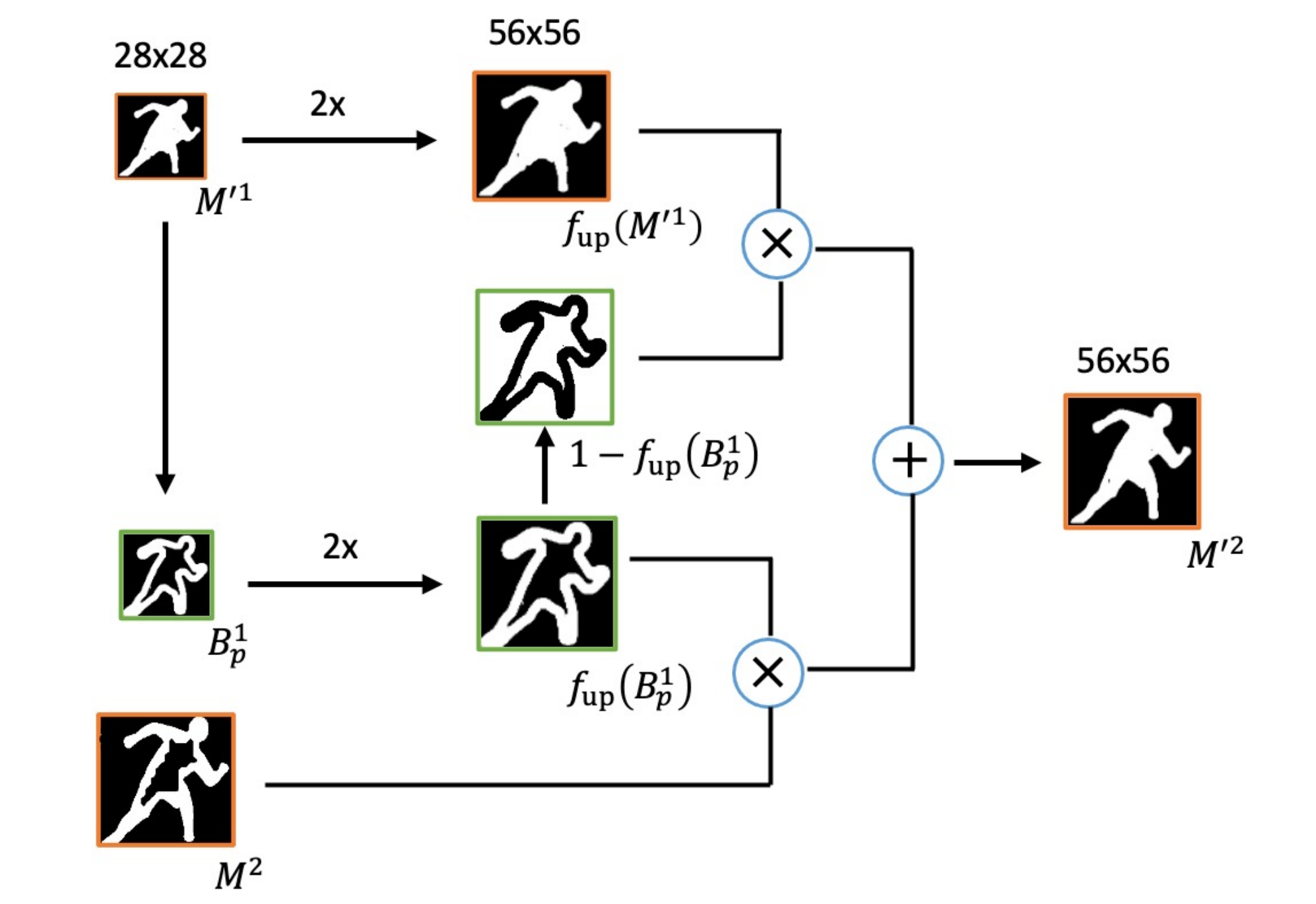}
        }
    \end{center}
    \caption{\textbf{The inference process of boundary-aware refinement (the second stage)}. $M'^1$ and $M^2$ are inputs of the BAR module, $B_p^1$ is the boundary region of $M'^1$, and $M'^2$ is the output mask of the second stage. The two operators with $\times$ and + denote pixel-wise multiplication and pixel-wise addition respectively.}
    \label{fig:process_of_boundary_aware_refinement}\vspace{-2mm}
\end{figure}

\noindent\textbf{Inference.} \label{inference_of_boundary_aware_refinement}For each instance, the first stage outputs a coarse and complete mask $M^{1}$ with size 28$\times$28, also generating its boundary mask $B_P^{1}$ (omitted in Figure~\ref{fig:framework}). The rule to generate the finer and complete instance mask $M'^k$ (the final output of stage $k$) can be formulated as follows:
\begin{small}
\begin{equation}
    M'^{1} = M^{1}
\end{equation}
\begin{equation}
    M'^{k} = f_{\text{up}}(B_P^{k-1})\otimes M^{k} + (1-f_{\text{up}}(B_P^{k-1})) \otimes f_{\text{up}}(M'^{k-1})
\end{equation}
\end{small}

\noindent where $\otimes$ denotes pixel-wise multiplication. Figure~\ref{fig:process_of_boundary_aware_refinement} displays the inference process of the second stage. We repeat this process until getting the finest mask.

\subsection{Implementation Details}
We adopted Mask R-CNN~\cite{MaskRCNN} as our baseline, and replaced the default mask head with our proposed multi-stage refinement head, and there were three refinement stages in the mask head by default.

All hyper-parameters were kept the same as Mask R-CNN implemented in MMDetection~\cite{mmdetection} except the new designed parts. We used the loss defined in Equation~\ref{boundary_loss} for the last two refinement stages. For the semantic head and the other mask prediction stages, we used the average binary cross-entropy loss. Losses for the other parts, including the RPN and the detection head, were kept the same as Mask R-CNN. Loss weights for the initial mask prediction stage and the three refinement stages were set as 0.25, 0.5, 0.75 and 1.0 respectively. To balance the losses between the detection head and the mask head, the loss weight for the detection head was set as 2.0, including both the classification and regression loss. $\hat{d}$ was set as 2 for training and 1 for inference. In addition, all models presented in the ablation experiments were trained with 1$\times$ learning schedule. No data augmentations except standard horizontal flipping were used unless otherwise stated.

\section{Experiments}
We performed extensive experiments on three standard instance segmentation datasets: COCO~\cite{COCO}, LVIS~\cite{LVIS} and Cityscapes~\cite{Cityscapes}. For all three datasets, we used the standard mask AP metric~\cite{COCO} as the evaluation metric.

\vspace{1mm}
\noindent \textbf{COCO} has 80 categories with instance-level annotations. Our models were trained on \textit{train2017}. Following~\cite{PointRend}, we also report AP$^\star$, which evaluates the COCO categories using LVIS annotations. As the LVIS annotations have significantly higher quality than COCO, it can better reflect improvements in mask quality. Note that the results for AP$^\star$ were from the same models trained on COCO.
\vspace{1mm}

\noindent \textbf{LVIS} is a long-tail instance segmentation dataset consisting of 1203 categories, having more than 2 million high-quality instance mask annotations. It contains about 100k, 20k, 20k images for training, validation and test respectively.
\vspace{1mm}

\noindent \textbf{Cityscapes} is a real-world dataset that consists of 2975, 500, 1525 images with resolution of 2048$\times$1024 for training, validation and test respectively. It contains 8 semantic categories for instance segmentation task.

\begin{table*}[t]
    \setlength{\abovecaptionskip}{-0.1cm}
    \begin{center}
    \scalebox{0.9}{
        \begin{tabular}{l|c|c|c|c|ccc}
        Method & Backbone & Schedule & AP & AP$^\star$ & AP$^\star_{S}$ & AP$^\star_{M}$ & AP$^\star_{L}$\\
        \thickhline
        Mask R-CNN          & R50-FPN    & 1$\times$ & 34.7 & 36.8 & 22.6 & 43.7 & 52.0 \\
        RefineMask           & R50-FPN    & 1$\times$ & 37.3 & 40.9 & 24.1 & 48.8 & 58.0 \\
        \hline
        Mask R-CNN          & R50-FPN    & 2$\times$ & 35.4 & 37.7 & 22.8 & 44.7 & 53.8 \\
        RefineMask        & R50-FPN    & 2$\times$ & 37.9 & 41.5 & 24.5 & 48.7 & 59.8  \\
        \hline
        Mask R-CNN          & R101-FPN   & 1$\times$ & 36.1 & 38.4 & 22.8 & 46.0 & 54.6 \\
        RefineMask        & R101-FPN   & 1$\times$ & 38.6 & 41.7 & 24.9 & 49.5 & 59.9 \\
        \hline
        Mask R-CNN          & R101-FPN   & 2$\times$ & 36.6 & 39.3 & 23.5 & 46.8 & 56.6 \\
        RefineMask        & R101-FPN   & 2$\times$ & 38.8 & 42.3 & 24.7 & 50.2 & 61.7 \\
        \hline
        Mask R-CNN$^\ddag$          & X101-FPN   & 3$\times$ & 39.4 & 41.8 & 27.2 & 49.0 & 57.7 \\
        RefineMask$^\ddag$        & X101-FPN   & 3$\times$ & 41.5 & 45.3 & 28.6 & 53.1 & 62.8 \\
        \end{tabular}
    }
    \end{center}
    \caption{\textbf{Comparison with Mask R-CNN on COCO \textit{val2017}.} Models with $^\ddag$ were trained with 3$\times$ schedule using multi-scale training with shorter side range [640, 800].}
    \label{validation_result}\vspace{-2mm}
\end{table*}

\subsection{Main Results}
We first evaluated our RefineMask with different backbones and different learning schedules on COCO \textit{val2017} (Table~\ref{validation_result}). RefineMask outperformed Mask R-CNN by a large margin under various configurations. Without bells and whistles, RefineMask achieved 2.6 points AP improvements over the Mask R-CNN baseline using ResNet-50-FPN as the backbone network. When evaluating the same COCO results using the 80 categories subset of LVIS annotations, RefineMask surpassed Mask R-CNN by 4.1 points AP$^\star$. This better demonstrates the effectiveness of RefineMask in predicting high-quality instance masks.

We also present runtime of RefineMask in Table~\ref{speed_comparison}. Compared to Mask R-CNN, RefineMask achieved significant improvement at a small amount of extra computational cost. Comparisons with PointRend~\cite{PointRend} and HTC~\cite{HTC} showed the superiority of RefineMask in both accuracy and speed.

\begin{table}[t]
    \setlength{\abovecaptionskip}{-0.1cm}
    \begin{center}
    \scalebox{0.9}{
        \begin{tabular}{c|c|c|c}
        Model & AP & AP$^\star$ & Runtime (fps) \\
        \thickhline
        Mask R-CNN & 34.7 & 36.8 & 15.7 \\
        PointRend & 35.6 & 38.7  & 11.4 \\
        HTC & 37.4 & 40.7 & 4.4 \\
        \textbf{RefineMask} & 37.3 & 40.9 & 11.4 \\
        \end{tabular}
    }
    \end{center}
    \caption{\textbf{Efficiency of RefineMask.}
    We reimplemented PointRend~\cite{PointRend} and HTC~\cite{HTC} for fair comparison. All models were trained with 1$\times$ schedule using R50-FPN as the backbone network. The inference time was measured on a single Tesla V100 GPU.}
    \label{speed_comparison}
\end{table}

\subsection{Ablations Experiments} \label{ablation_study}
We conducted extensive ablation experiments on COCO \textit{val2017} to analyze RefineMask. \vspace{2mm}

\noindent\textbf{Different number of stages.}
We compared models with different number of stages. For each additional stage, the output size is twice larger than its preceding stage. The results are shown in Table~\ref{train_with_multi_stage}. Models with more stages (also larger output size) obtained significant performance improvements and the large objects benefited most from that. The model with three stages and the model with four stages obtained comparable performance, but the former one obviously had lower computational cost. \vspace{2mm}

\noindent\textbf{Effectiveness of SFM.} We compared the SFM with other three fusion modules (Table \ref{fusion_module_comparison}). Our proposed SFM performed much better than the other three fusion modules, as it has the largest receptive field and can better capture the context information for each single neuron. \vspace{2mm}

\begin{table}[t]
    \setlength{\abovecaptionskip}{-0.2cm}
    \begin{center}
        \scalebox{0.9}{
            \begin{tabular}{c|c|c|c|ccc}
            Stages & Output size & AP & AP$^\star$ & AP$^\star_S$ & AP$^\star_M$ & AP$^\star_L$\\
            \thickhline
            1 & 28$\times$28  & 35.7 & 38.3 &  23.1 & 45.6 & 54.4 \\
            2 & 56$\times$56  & 36.6 & 40.3 &  23.3 & 47.9 & 57.3 \\
            3 & 112$\times$112& \textbf{37.3} & 40.9 &  24.1 & \textbf{48.8} & 58.0 \\
            4 & 224$\times$224& 37.1 & \textbf{41.0} &  \textbf{24.3} & 48.5 & \textbf{58.7} \\
            \end{tabular}
        }
    \end{center}
    \caption{\textbf{Different number of stages.} Model with more refinement stages has larger output size.}
    \label{train_with_multi_stage}\vspace{-2mm}
\end{table}

\noindent\textbf{Effectiveness of fine-grained features.}
We analyzed effectiveness of the fine-grained features by removing the semantic head in RefineMask. The results are shown in Table~\ref{effectiveness_of_fine-grained_features}. With fine-grained features, RefineMask brought an improvement of 1.6 points AP and 2.5 points AP$^\star$, which indicates the usefulness of fine-grained features for high-quality mask prediction. In addition, objects with large scales benefited more from fine-grained features, indicated by the large AP$^\star_L$ improvement (54.3$\to$58.0).\vspace{2mm}

\begin{table}[t]
    \setlength{\abovecaptionskip}{-0.2cm}
    \begin{center}
    \scalebox{0.9}{
        \begin{tabular}{c|c|c|ccc}
        Fusion module & AP & AP$^\star$ & AP$^\star_S$ & AP$^\star_M$ & AP$^\star_L$\\
        \thickhline
        1 single 1$\times$1 Conv  & 35.8 & 38.7 & 22.5 & 46.1 & 55.3 \\
        1 single 3$\times$3 Conv  & 36.3 & 39.9 & 23.4 & 47.1 & 57.2 \\
        3 parallel 3$\times$3 Convs  & 36.7 & 39.9 & 23.4 & 47.2 & 57.6 \\
        \textbf{SFM} & \textbf{37.3} & \textbf{40.9} &  \textbf{24.1} & \textbf{48.8} & \textbf{58.0}
        \end{tabular}
    }
    \end{center}
    \caption{\textbf{Different fusion modules.} Comparison of different designs of fusion module.}
    \label{fusion_module_comparison}
\end{table}

\begin{table}[t]
    \setlength{\abovecaptionskip}{-0.2cm}
    \begin{center}
        \scalebox{0.9}{
            \begin{tabular}{c|c|c|ccc}
            Semantic head & AP & AP$^\star$ & AP$^\star_S$ & AP$^\star_M$ & AP$^\star_L$\\
            \thickhline
            & 35.7 & 38.4 & 23.1 & 45.7 & 54.3 \\
            \checkmark & \textbf{37.3} & \textbf{40.9} & \textbf{24.1} & \textbf{48.8} & \textbf{58.0} \\
            \end{tabular}
        }
        \end{center}
    \caption{\textbf{Effectiveness of fine-grained features.} Without the semantic head, the mask head only relies on the instance features to predict instance masks.}
    \label{effectiveness_of_fine-grained_features}
\end{table}

\begin{table}[t]
    \setlength{\abovecaptionskip}{-0.2cm}
    \begin{center}
        \scalebox{0.9}{
            \begin{tabular}{c|c|c|ccc}
            Multi-stage refinement & AP & AP$^\star$ & AP$^\star_S$ & AP$^\star_M$ & AP$^\star_L$\\
            \thickhline
            & 36.3 & 39.5 & 23.1 & 46.9 & 56.2 \\
            \checkmark & \textbf{37.3} & \textbf{40.9} & \textbf{24.1} & \textbf{48.8} & \textbf{58.0} \\
            \end{tabular}
        }
        \end{center}
    \caption{\textbf{Effectiveness of multi-stage refinement.} Only the last stage was supervised without multi-stage refinement.}
    \label{effectiveness_of_MSR}
\end{table}

\begin{table}[t]
    \setlength{\abovecaptionskip}{-0.2cm}
    \begin{center}
        \scalebox{0.9}{
            \begin{tabular}{c|c|c|ccc}
            BAR & AP & AP$^\star$ & AP$^\star_S$ & AP$^\star_M$ & AP$^\star_L$\\
            \thickhline
            & 36.9 & 40.0 &  23.6 & 47.4 & 56.8 \\
            \checkmark & \textbf{37.3} & \textbf{40.9} & \textbf{24.1} & \textbf{48.8} & \textbf{58.0} \\
            \end{tabular}
        }
        \end{center}
    \caption{\textbf{Effectiveness of boundary-aware refinement.} BAR denotes boundary-aware refinement. Each stage predicted a complete mask for each instance without boundary-aware refinement.}
    \label{effectiveness_of_BAR}\vspace{-2mm}
\end{table}

\begin{table*}[t]
    \setlength{\abovecaptionskip}{-0.1cm}
    \begin{center}
    \scalebox{0.9}{
        \begin{tabular}{l|c|c|c|ccc|c}
        Method & Backbone & AP$_{dev}$ & AP$^\star$ & AP$^\star_{S}$ & AP$^\star_{M}$ & AP$^\star_{L}$ & Runtime (fps)\\
        \thickhline
        BMask R-CNN~\cite{BMaskRCNN} &           & 35.9 & - & - & - & - & -\\
        Mask R-CNN$^\dag$          &            & 35.7 & 37.7 & 22.8 & 44.7 & 53.8 & 15.7 \\
        PointRend$^\dag$           &     & 36.8 & 39.7 & 22.9 & 46.7 & 57.5 & 11.4 \\
        HTC~\cite{HTC}      &     R50-FPN & 38.4 & - & - & - & - & 4.4 \\
        HTC$^\dag$      &            & 38.6 & 41.3 & 25.0 & 48.5 & 58.8 & 4.4 \\
        \textbf{RefineMask}          &           & 38.2 & 41.5 & 24.5 & 48.7 & 59.8 & 11.4 \\
        \textbf{RefineMask}$^\ddag$   &          & \textbf{40.2} & \textbf{43.4} & \textbf{27.5} & \textbf{50.6} & \textbf{60.7} & 11.4 \\
        \hline
        Mask R-CNN~\cite{MaskRCNN}   &           & 35.7 & - & - & - & - & - \\
        BMask R-CNN~\cite{BMaskRCNN} &           & 37.7 & - & - & - & - & -\\
        Mask R-CNN$^\dag$           &           & 37.1 & 39.3 & 23.5 & 46.8 & 56.6 & 13.5 \\
        PointRend$^\dag$            &   & 38.2 & 41.4 & 24.7 & 49.0 & 59.8 & 10.0\\
        HTC~\cite{HTC}              &  R101-FPN  & 39.7 & - & - & - & - & 3.9 \\
        HTC$^\dag$                  &        & 39.7 & 42.5 & 26.2 & 50.4 & 60.4 & 3.9 \\
        \textbf{RefineMask}           &           & 39.4 & 42.3 & 24.7 & 50.2 & 61.7 & 9.6 \\
        \textbf{RefineMask}$^\ddag$   &           & \textbf{41.2} & \textbf{44.6} & \textbf{27.7} & \textbf{52.5} & \textbf{63.2} & 9.6 \\
        \hline
        Mask R-CNN~\cite{MaskRCNN}   &           & 37.1 & - & - & - & - & - \\
        Mask R-CNN$^\dag$$^\ddag$    &           & 39.6 & 41.8 & 27.2 & 49.0 & 57.7 & 8.0\\
        PointRend$^\dag$$^\ddag$ &   & 41.1 & 44.4 & 27.8 & 52.0 & 62.0 & 6.7 \\
        HTC~\cite{HTC}  &  X101-FPN & 41.2 & - & - & - & - & 3.4 \\
        HTC$^\dag$  &        & 41.3 & 44.1 & 27.2 & 51.9 & 61.5 & 3.4 \\
        \textbf{RefineMask}            &           & 41.0 & 43.6 & 25.8 & 51.8 & 62.2 & 6.6 \\
        \textbf{RefineMask}$^\ddag$    &           & \textbf{41.8} & \textbf{45.3} & \textbf{28.6} & \textbf{53.1} & \textbf{62.8} & 6.6 \\
        \end{tabular}
    }
    \end{center}
    \caption{\textbf{Comparisons of single-model results on COCO \textit{val2017} and \textit{test-dev}.} $^\dag$ denotes our implementation. AP$_{dev}$ denotes the evaluation results on \textit{test-dev}, and AP$^\star$ denotes the evaluation results on COCO \textit{val2017} using the 80 categories subset of LVIS. Note that HTC employs the same detection head as Cascade R-CNN ~\cite{CascadeRCNN} and uses the extra COCO-Stuff~\cite{COCOStuff} annotations. Models with $^\ddag$ were trained with 3$\times$ schedule using multi-scale training with shorter side range [640, 800].}
    \label{coco_test_dev}\vspace{-2mm}
\end{table*}

\noindent\textbf{Effectiveness of multi-stage refinement.}
We designed an ablation experiment to prove the necessity of multi-stage refinement by removing supervision of all previous stages. The result is shown in Table~\ref{effectiveness_of_MSR}. With multi-stage refinement, RefineMask improved the AP$^\star$ by 1.4 points and the AP$^\star_L$ by 1.8 points, indicating that multi-stage refinement is important to predict high-quality instance masks.\vspace{2mm}

\noindent\textbf{Effectiveness of boundary-aware refinement.}
We also conducted an ablation experiment to analyze effectiveness of the boundary-aware refinement. Without boundary-aware refinement, each stage of RefineMask predicts complete masks. Here we only used the output from the last stage, which had the largest output size. The results are shown in Table~\ref{effectiveness_of_BAR}. With boundary-aware refinement, RefineMask improved the AP$^\star$ by 0.9 points.
Although the output size becomes higher in later stages, pixels far away from the object contour cannot benefit more from this, as these pixels generally belongs to the easy non-boundary regions.

\begin{figure*}[t]
    \begin{center}
        \includegraphics[width=1.0\linewidth]{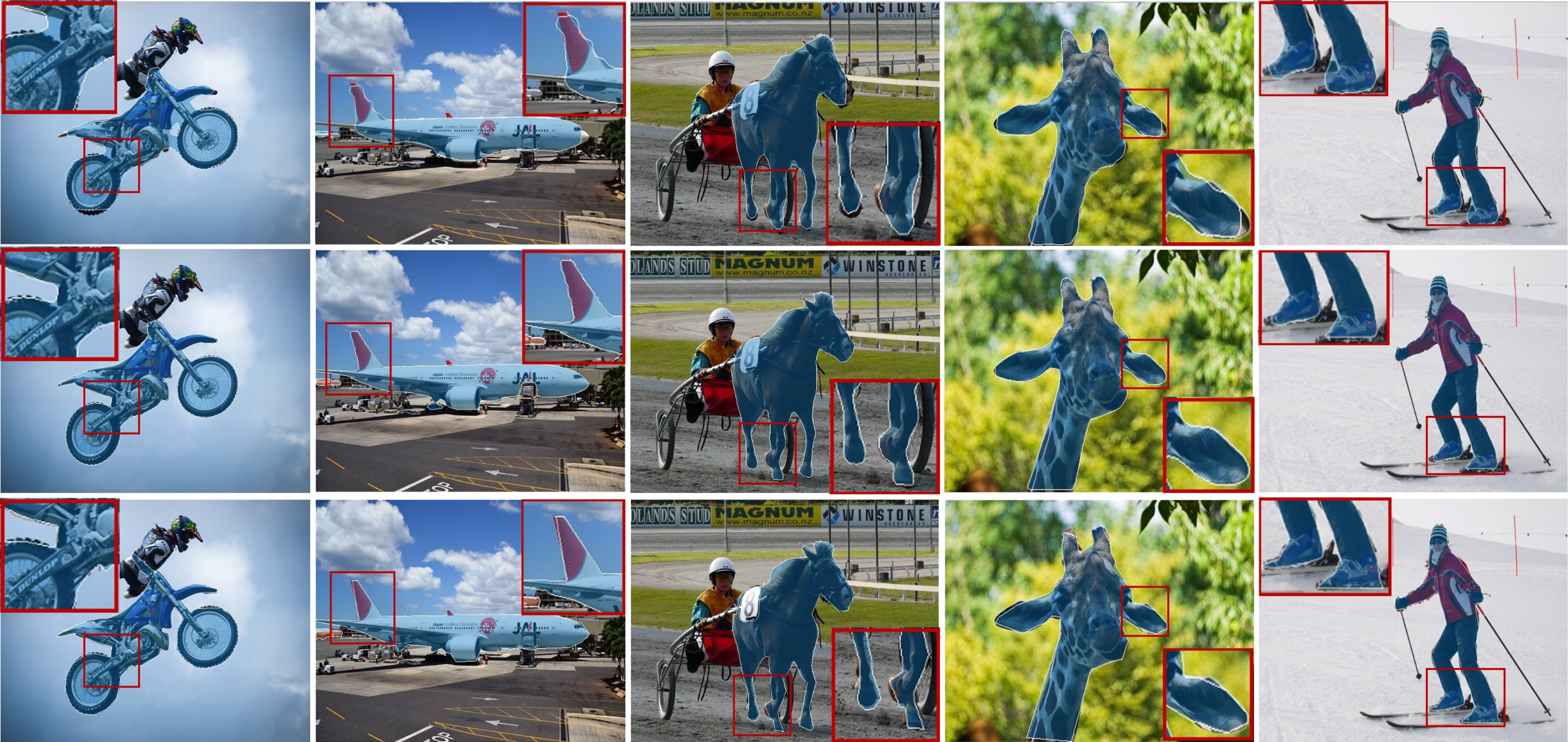}
    \end{center}
    \caption{\textbf{COCO example result tuples from Mask R-CNN~\cite{MaskRCNN} (the top row), RefineMask (the middle row) and the ground truth masks (the bottom row).} RefineMask predicted masks with substantially higher quality than Mask R-CNN, even better than the ground truth masks, especially on the sharp boundaries.}
    \label{fig:visualization_results}\vspace{-3mm}
\end{figure*}

\subsection{Comparison with Previous Methods}
We present single-model results on both COCO \textit{val2017} (AP$^\star$) and \textit{test-dev} (AP$_{dev}$) in Table~\ref{coco_test_dev} to compare RefineMask with previous methods. Compared with Mask R-CNN, RefineMask improved the performance by 3.8, 3.0, 3.5 points AP$^\star$ with ResNet-50-FPN, ResNet-101-FPN and ResNext-101-FPN as the backbone network respectively. Without utilizing the more powerful detection head in Cascade R-CNN~\cite{CascadeRCNN} and the extra COCO-Stuff~\cite{COCOStuff} data annotations, RefineMask still achieved comparable performance as HTC, with a much faster inference time. RefineMask also outperformed the PointRend~\cite{PointRend} under various configurations with comparable speed. As \cite{PointRend} did not release the experimental results on COCO \textit{test-dev}, we reimplemented it following the same configurations except that we used the pytorch-style ImageNet~\cite{ImageNet} pretrained models and did not use multi-scale data augmentation during training for fair
comparison unless otherwise stated. Our reimplemented version had similar performance with~\cite{PointRend}.

\begin{table}[t]
    \setlength{\abovecaptionskip}{-0.1cm}
    \begin{center}
    \scalebox{0.9}{
        \begin{tabular}{l|l|l|c|c|c}
        Method & Backbone & AP & AP$_{r}$ & AP$_{c}$ & AP$_{f}$\\
        \thickhline
        Mask R-CNN  & R50-FPN & 22.1 & 10.1 & 21.7 & 30.0 \\
        \textbf{RefineMask} & R50-FPN & \textbf{25.5} & \textbf{14.2} & \textbf{24.3} & \textbf{31.7} \\
        \hline
        \multicolumn{2}{c|}{}& \textbf{+3.4} & \textbf{+4.1} & \textbf{+2.6} & \textbf{+1.7}
        \end{tabular}
    }
    \end{center}
    \caption{\textbf{Results on LVISv1.0 validation set.} All Models were trained with 1$\times$ schedule and the hyper-parameters were kept the same as~\cite{LVIS}.}
    \label{LVIS_validation}\vspace{-2mm}
\end{table}

\subsection{Experiments on LVIS}
We first evaluated our RefineMask on the LVIS validation set. To save training memory, we replaced the default class-specific classifier in the last stage with a class-agnostic classifier. The results are shown in Table~\ref{LVIS_validation}. RefineMask achieved 3.4 points AP improvement compared with the Mask R-CNN baseline, which is larger than that on COCO due to the finer annotations.

We then managed to obtain a good result on the LVIS \textit{test-dev} dataset based on RefineMask. Specifically, we combined the mask head of the model presented in~\cite{LVIS2020Winner} with our multi-stage refinement head. Following the rule of the LVIS Challenge 2020, we did not use any extra data with human labels. Our final single-model result outperformed the winner of LVIS challenge 2020 by 1.3 points (Table~\ref{LVIS_test_dev}).

\begin{table}[t]
    \setlength{\abovecaptionskip}{-0.1cm}
    \begin{center}
    \scalebox{0.9}{
        \begin{tabular}{l|c|c|c|c}
        Method & AP & AP$_r$ & AP$_c$ & AP$_f$\\
        \thickhline
        lvisTraveler~\cite{LVIS2020Winner}  & 41.2 & 31.9 & 40.4 & 46.4 \\
        lvisTraveler + \textbf{RefineMask} & \textbf{42.5} & \textbf{33.5} & \textbf{41.5} & \textbf{47.7} \\
        \hline
        & \textbf{+1.3} & \textbf{+1.6} & \textbf{+1.1} & \textbf{+1.3}
        \end{tabular}
    }
    \end{center}
    \caption{\textbf{Results on LVISv1.0 test-dev set.} The first row is result of the winner of the LVIS challenge 2020.}
    \label{LVIS_test_dev}
\end{table}

\begin{table}[t]
    \setlength{\abovecaptionskip}{-0.1cm}
    \begin{center}
    \scalebox{0.9}{
        \begin{tabular}{l|l|l|c|c|c}
        Method & Backbone & AP & AP$_{S}$ & AP$_{M}$ & AP$_{L}$\\
        \thickhline
        PointRend~\cite{PointRend} & R50-FPN & 35.8 & - & - & - \\
        Mask R-CNN  & R50-FPN & 33.8 & 12.0 & 31.5 & 51.8 \\
        \textbf{RefineMask} & R50-FPN & \textbf{37.6} & \textbf{14.6} & \textbf{34.0} & \textbf{58.1} \\
        \hline
        \multicolumn{2}{c|}{} & \textbf{+3.8} & \textbf{+2.6} & \textbf{+2.5} & \textbf{+6.3}
        \end{tabular}
    }
    \end{center}
    \caption{\textbf{Results on Cityscapes validation set.} All models were trained on the fine annotations with 64 epochs, using multi-scale training with shorter side range [800, 1024].}
    \label{cityscapes_validation}\vspace{-2mm}
\end{table}

\subsection{Experiments on Cityscapes}
We also evaluated RefineMask on Cityscapes (Table~\ref{cityscapes_validation}). As the experimental results on Cityscapes have high variance, we report the median of three runs. RefineMask obtained larger improvements on this dataset than on both COCO and LVIS. This further demonstrates the superiority of our approach for high-quality mask prediction.

\subsection{Qualitative results.}
We show some visualization examples from COCO in Figure~\ref{fig:visualization_results}. RefineMask predicted masks with substantially higher quality than Mask R-CNN, especially for the hard regions, such as  the tail of the airplane (the second column), the feet of the horse (the third column), and so on. As the COCO annotations are much coarser than LVIS and Cityscapes, RefineMask even yielded more accurate masks than the ground truth masks, which further indicates the necessity of a more stringent evaluation metric (AP$^\star$).

\section{Conclusion}
In this work, we propose a multi-stage framework called RefineMask towards high-quality instance segmentation. Previous two stage methods are struggling to predict accurate masks, as they depend on the pooling operation, \eg RoIAlign, to extract instance features from the feature pyramid, which loses details required for predicting crisp boundaries. To alleviate this problem, RefineMask refines instance masks by incorporating fine-grained features iteratively during the instance-wise segmentation process. We believe RefineMask can serve as a strong baseline for high-quality instance segmentation.\vspace{2mm}

\noindent\textbf{Acknowledgement.} This work was supported in part by the National Key Research and Development Program of China (No. 2017YFA0700904), and the National Natural Science Foundation of China (Nos. 61836014, U19B2034, 62061136001 and 61620106010).

\newpage

\normalem

{\small
\bibliographystyle{ieee_fullname}
\bibliography{egbib}
}

\vspace{20mm}
\section*{Appendix}

\noindent \subsection*{A.1 Boundary Region Calculation}
As mentioned in Section~\ref{boudare_aware_refinement}, we design a convolutional operator to approximate the calculation of boundary regions defined in Equation~\ref{boundary_definition} for efficient implementation. In this section, we give details of the convolutional operator. \vspace{2mm}

\noindent\textbf{Implementation of the approximation approach.}
To calculate the boundary region $B^k$ of instance mask $M^k$, we use a specific convolutional operator to calculate the foreground boundary region and the background boundary region respectively, where $k$ denotes index of refinement stages in the mask head. When calculating the background boundary region, we first reverse binary values of $M^k$. Taking boundary width of 1 as an example, kernel weights of the operator are defined as a 3$\times$3 matrix:

\begin{equation}\nonumber
\left[
    \begin{matrix}
        -1 & -1 & -1 \\
        -1 &  8 & -1 \\
        -1 & -1 & -1
        \end{matrix}
\right]
\end{equation}

\noindent Let $D^{k}$ denotes the intermediate output of above convolutional layer, and it has the same size as $M^{k}$. Values of $B^k$ are determined as follows:
\begin{equation}\nonumber
    B^k(i,j) = \left\{
            \begin{array}{lr}
            1, & \text{if } D^k(i,j) > 0 \\
            0, & \text{otherwise.}\\
            \end{array}
        \right.
\end{equation}
Specifically, the custom operator defined above executes on a binary instance mask $M^k$, and it generates the binary boundary mask $B^k$ of $M^k$. If the operator calculates on a pure foreground region (all ones) or background region (all zeros), the output $D^k(i,j)$ is zero. Only when the operator calculates on a boundary region, the output $D^k(i,j)$ can be larger than zero. To calculate boundary regions with width of 2, the operator can be defined as a 5$\times$5 matrix:
\begin{equation}\nonumber
    \left[
        \begin{matrix}
            -1 & -1 & -1 & -1 & -1 \\
            -1 & -1 & -1 & -1 & -1 \\
            -1 & -1 & 24 & -1 & -1 \\
            -1 & -1 & -1 & -1 & -1 \\
            -1 & -1 & -1 & -1 & -1 \\
            \end{matrix}
    \right]
\end{equation}
To calculate boundary regions with larger width, the convolutional operator can be defined similarly, making sure sum of all weights is zero.\vspace{2mm}

\begin{table}[t]
    \begin{center}
    \scalebox{0.9}{
        \begin{tabular}{c|c}
        Output size & IoU \\
        \thickhline
        28$\times$28 & 0.76 \\
        56$\times$56 & 0.75 \\
        112$\times$112 & 0.80 \\
        \end{tabular}
    }
    \end{center}\
    \caption{IoU between boundary regions from two implementations.}
    \label{IoU_comparision}
\end{table}

\begin{table}[t]
    \begin{center}
    \scalebox{0.9}{
        \begin{tabular}{c|c|c|c|c|c}
        Boundary width & AP & AP$^\star$ & AP$^\star_{S}$ & AP$^\star_{M}$ & AP$^\star_{L}$\\
        \thickhline
        1 & 37.2 & 40.7 & 24.1 & 47.7 & 57.7 \\
        2 & 37.3 & 40.9 & 24.1 & 48.8 & 58.0 \\
        3 & 37.3 & 40.8 & 23.6 & 48.4 & 58.4 \\
        \end{tabular}
    }
    \end{center}
    \caption{Different boundary widths for boundary-aware refinement.}
    \label{Different_boundary_widths}
\end{table}

\begin{table}[t]
    \begin{center}
        \scalebox{0.9}{
            \begin{tabular}{c|c|c|c|c|c}
            Model & Backbone & AP & AP$^\star$ & F$_{1px}$ & F$_{3px}$ \\
            \thickhline
            Mask R-CNN & X101-FPN & \textbf{37.8} & 40.1 & 64.1 & 82.6\\
            Mask R-CNN & R50-FPN & 34.7 & 36.8 & 62.0 & 80.6 \\
            \textbf{RefineMask} & R50-FPN & 37.3 & \textbf{40.9} & \textbf{69.6} & \textbf{84.9} \\
            \hline
            \multicolumn{2}{c|}{} & \textbf{+2.6} & \textbf{+4.1} & \textbf{+7.6} & \textbf{+4.3}
            \end{tabular}
        }
    \end{center}
    \caption{Comparison between Mask R-CNN and RefineMask on COCO $val2017$ under different evaluation metrics.}
    \label{AF_Metric}
 \end{table}

\begin{figure*}[t]
    \begin{center}
    \includegraphics[width=1.0\textwidth]{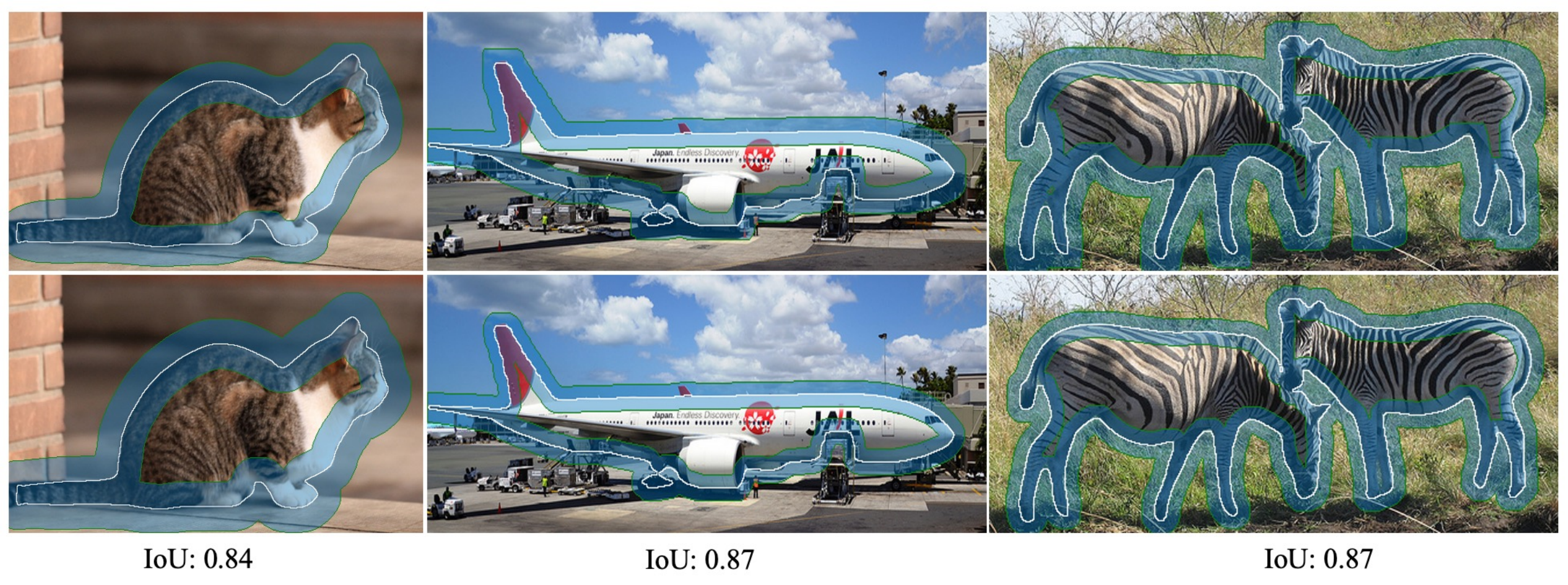}
    \end{center}
    \caption{Visualization results of the approximated boundary regions by convolutional kernel (the top row) and the defined boundary regions by Equation~\ref{boundary_definition} in Section~\ref{boudare_aware_refinement} (the bottom row).}
    \label{fig:boundary_comparison}
\end{figure*}

\noindent\textbf{Comparisons between two implementations.} We counted the average IoU between boundary regions generated by the definition and the approximation approach (Table~\ref{IoU_comparision}). The high IoU indicates our faster implementation generates similar results with the definition. We also give examples of boundary regions calculated by these two methods in Figure~{\ref{fig:boundary_comparison}} respectively. As we can see, boundary regions generated by these two methods are visually near the same.
In addition, experiments show that the effectiveness of the boundary-aware refinement is not sensitive to the boundary width (Table~\ref{Different_boundary_widths}), which further indicates the subtle differences between these two implementations are not essential to the final performance (the differences between boundary regions with different boundary widths are obviously much larger than the differences between results generated by these two implementations).

\noindent \subsection*{A.2 Direct Measurement of Boundary Quality}
In order to directly measure the boundary quality, we also evaluated our method by the metric F$_{npx}$, which is designed by adapting the boundary F1 score proposed in \cite{DAVISCVPR} from semantic segmentation to instance segmentation. For each instance, we computed the boundary F1 score within $n$ pixels from object contour between the ground truth mask and its positive prediction mask with maximum IoU. It was ignored if there was no positive prediction for a given instance. Results in Table~\ref{AF_Metric} show that more gains are observed on COCO, where refinemask even outperforms the Mask R-CNN model with heavier backbone X101-FPN by a large margin, further indicating that RefineMask improves the boundary quality significantly.

\newpage

\color{white}{hidden fonts}

\end{document}